\documentclass[letterpaper, 10 pt, conference]{ieeeconf}  % Comment this line out if you need a4paper

\IEEEoverridecommandlockouts                              % This command is only needed if 
\usepackage{graphicx}
\usepackage{adjustbox}
\usepackage{multicol}
\usepackage{tabularx}
\usepackage{caption}
\usepackage{amssymb}
\usepackage{cellspace}
\usepackage{hyperref}
\usepackage{cite}
\usepackage[table]{xcolor}
\usepackage{multirow}
\title{\LARGE \bf
SCARP: 3D Shape Completion in ARbitrary Poses for Improved Grasping
}

\author{Bipasha Sen$^{*1}$, Aditya Agarwal$^{*1}$, Gaurav Singh$^{*1}$, Brojeshwar B.$^{2}$, Srinath Sridhar$^{3}$, and Madhava Krishna$^{1}$% <-this % stops a space
\thanks{$^*$Equal Contributions}% <-this % stops a space
\thanks{$^{1}$Robotics Research Center, IIIT-Hyderabad}%
\thanks{$^{2}$TCS Research, India}%
\thanks{$^{3}$Brown University}%
}

\begin{document}

\maketitle

\thispagestyle{empty}
\pagestyle{empty}

%%%%%%%%%%%%%%%%%%%%%%%%%%%%%%%%%%%%%%%%%%%%%%%%%%%%%%%%%%%%%%%%%%%%%%%%%%%%%%%%
\begin{abstract}
Recovering full 3D shapes from partial observations is a challenging task that has been extensively addressed in the computer vision community. Many deep learning methods tackle this problem by training 3D shape generation networks to learn a prior over the full 3D shapes. In this training regime, the methods expect the inputs to be in a fixed canonical form, without which they fail to learn a valid prior over the 3D shapes. 
% This limits their applications in many Robotics tasks where the partial observations are made in arbitrary poses \kd{rewrite}. 
We propose SCARP, a model that performs \underline{S}hape \underline{C}ompletion in \underline{AR}bitrary \underline{P}oses. Given a partial pointcloud of an object, SCARP learns a disentangled feature representation of pose and shape by relying on rotationally equivariant pose features and geometric shape features trained using a multi-tasking objective. Unlike existing methods that depend on an external canonicalization, SCARP performs canonicalization, pose estimation, and shape completion in a single network, improving the performance by 45\% over the existing baselines.
In this work, we use SCARP for improving grasp proposals on tabletop objects. By completing partial tabletop objects directly in their observed poses, SCARP enables a SOTA grasp proposal network improve their proposals by 71.2\% on partial shapes. {Project page: \textcolor{orange}{\href{https://bipashasen.github.io/scarp}{https://bipashasen.github.io/scarp}}}

\end{abstract}

%%%%%%%%%%%%%%%%%%%%%%%%%%%%%%%%%%%%%%%%%%%%%%%%%%%%%%%%%%%%%%%%%%%%%%%%%%%%%%%%
\section{INTRODUCTION}

% Features
% Canoncalization of partial objects (for a fixed frame)
% Estimation of E
% Symmetrical Objects
% 

% Validations needed
% Grasp loss
% Table-top and non-table top scenes.
% improved completion with improved number of views.
% Ablations - without pointnet, without shape completion how good is the canonicalization, Without the new proposed loss of direct correspondences how good is the canonicalization of the symmetrical objects?

Given a partial observation of an object, 3D shape completion aims to recover the full 3D shape of the object. This has been widely addressed in computer vision~\cite{shapeformer, mvp, autosdf, deepsdf, metasdf, seedformer, pointr, e2ecad} and has many diverse downstream applications in robotics including visual servoing~\cite{visualserv1}, manipulation~\cite{challenges2, ourpaper, 6dof, contact-graspnet}, visual inspection~\cite{inspection1}, autonomous driving~\cite{autonom1, autonom2, autonom3}. 

% \Srinath{I would suggest proposing a method for ``shape completetion'' where the representation is a point cloud. The method can potentially be extended to other shape representations as well, so it's worth saying so.}

% Pointclouds are the most common form of 3D representation~\cite{guo2020deep, s19194188}.
%widely used in autonomous driving (LiDAR Scans)~\cite{kitti}, robotic manipulation (captured using depth cameras)~\cite{ocrtoc, shapnet}. 
Many existing methods tackle shape completion by 
% Existing methods tackle partial pointcloud completion by
incorporating a training scheme that learns a prior over the full 3D shapes. This is done by training an autoencoder~\cite{crn, pcn, shapeformer, seedformer} or a GAN~\cite{sinv} over many different instances of full shapes. At inference, this learned prior space is conditionally queried on the partial observations. 
These methods however, suffer from a major limitation: they expect the partial input to be in a fixed canonical frame--a common frame of reference that is shared between instances in that category~\cite{nocs,condor}. 
A particular shape 
% A partial pointcloud\Srinath{shape} 
$X$ in two different poses $\{R_1, T_1\}$ and $\{R_2, T_2\}$ will have very different geometry. 
% \Srinath{not clear what `points' are referring to, explain}.  
% a pose $\{R, T\}$ will have very different point locations than $X_p$ in a different pose $\{R', T'\}$. 
As a result, $X$ in different poses appear as novel instances for these methods 
% any new pose is a novel instance for these methods 
inhibiting them from learning a valid prior over shapes. 

Existing datasets like ShapeNet~\cite{shapenet} have shapes that are manually aligned to a canonical frame, but real shape observations (e.g., depth maps) do not contain this information.
% A canonical frame is a category-level frame of reference that is consistent relative to the geometry of the shape under different 3D poses.
% This limits the applications of these models in the real life where the objects are observed in arbitrary poses.
% In the real-world, pointclouds are captured in arbitrary poses. For instance, in robotic manipulation, a 3D scene is reconstructed by capturing the scene using depth cameras from arbitrary camera poses. Similarly, in autonomous driving, LiDAR scans are captured from a moving frame of references. The objects in the captured scans (cars, benches, bottles, mugs, etc.) are thus observed in arbitrary poses. This is a challenge for the existing methods that cannot directly operate on arbitrarily positioned partial observations.
% \Srinath{The above para can be replaced with a single line and some refs.}
A naive approach to tackling this challenge is to \emph{canonicalize}, i.e.,~map a 3D (full or partial) shape to a category-level canonical frame with~\cite{nocs} or without supervision~\cite{condor, capsule_networks,compass}. 
%. Canonicalization of partial pointclouds have been tackled in recent years~\cite{condor, capsule_networks,compass}. 
A multi-stage pipeline can be built involving the sequential steps of (1) canonicalization, (2) shape completion, and (3) de-canonicalization (bringing the object back in the original pose). 
In such a pipeline however, the performance of a shape completion network directly depends on the output quality of the canonicalization module. This can lead to errors propagating between these modules leading to a sub-optimal completion. 
%This can lead to error propogation and a sub-optimal reconstruction. 

\begin{figure}[t]
    \centering    
    \includegraphics[width=\linewidth]{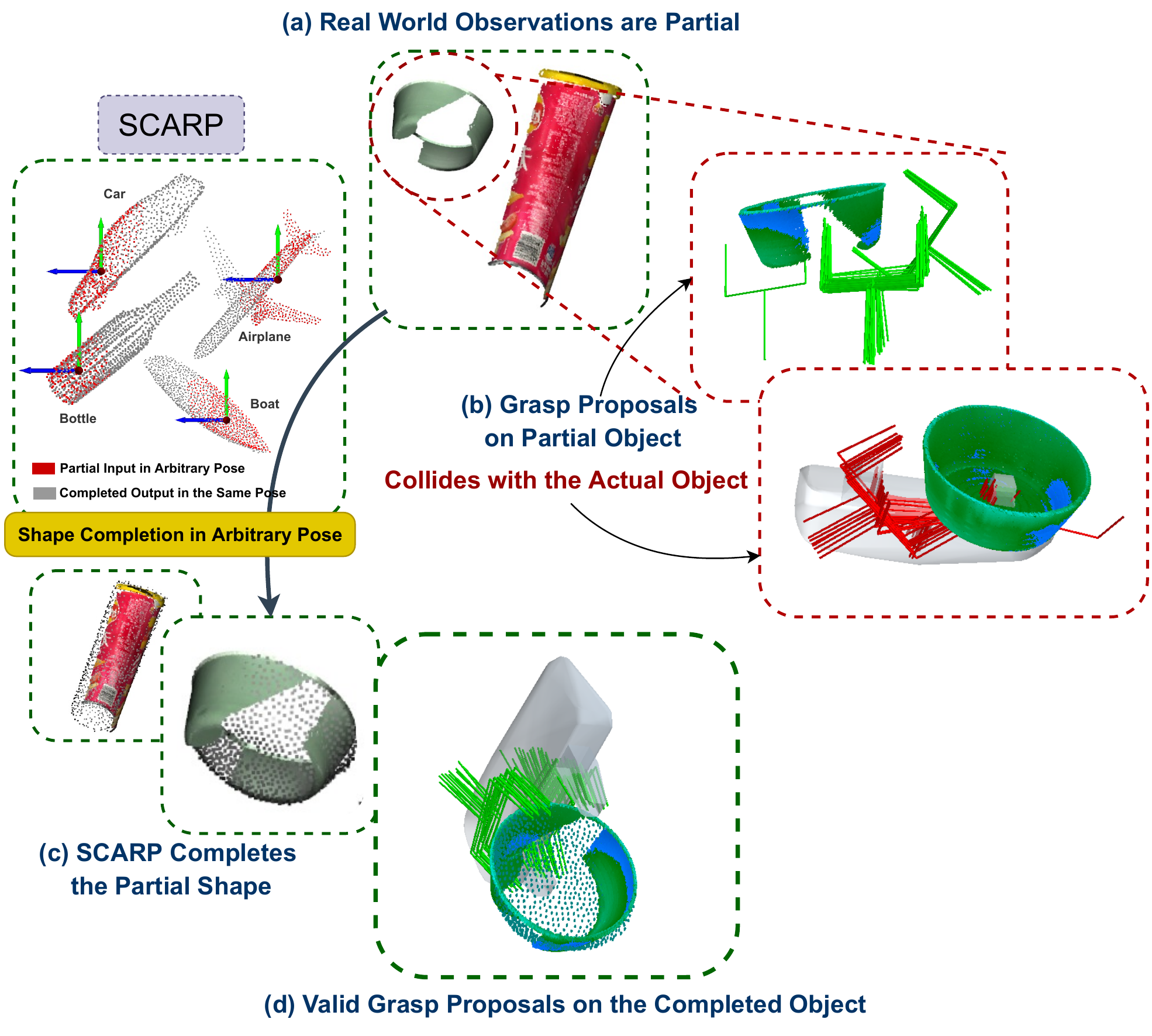}
    \caption{\small SCARP performs \underline{S}hape \underline{C}ompletion in \underline{AR}bitrary \underline{P}oses (top-left). We show an example of a real scene made of two tabletop objects. (a) The captured scene is partial leaving out a portion of the objects. (b) This results in grasp poses (in green) on the partial point cloud that directly collide (in red) with the actual object leading to a collision between the object and the Franka Panda's gripper (grey). (c) SCARP improves grasp proposal by accurately completing the partial pointcloud in the observed pose. (d) This enables the grasp proposal network to propose grasp poses (shown in green) on the completed pointcloud that do not collide with the actual object. }
    %The Franka's geripper is once again shown in grey}
    % Real world scene observations are incomplete and lead to grasp proposals that collide with the actual objects; SCARP improves the grasp proposal by performing accurate shape completion in the observed object pose.  }
    % (a) Capturing the scene from a particular view results in an incomplete pointcloud. (b) SOTA grasp proposal networks fail to estimate proper grasp poses in such partial views resulting in proposals that coincide with the actual object on the table. This results in a collision between the manipulator and the actual object. 
    % fail to estimate proper grasp poses and propose grasps in  in such partial views resulting in a collision between the manipulator and the actual object. 
    % (c) We use SCARP to complete the shapes of the two object directly in the given pose that results in an improved grasp pose estimation as shown in (d).}
    % For instance, grasp poses proposed on partial observation lead to a collision between the manipulator and the object (bottom-left). Accurate reconstructions using SCARP result in better grasp proposals and avoid such manipulator-object collisions (bottom-right).}
    \label{fig:banner}
\end{figure}

% \begin{figure}[t]
%     \centering    \includegraphics[width=\linewidth, height=3cm]{example-image} 
%     \caption{Grasp proposals using \cite{contact-graspnet}. Grasp proposals shown in red are coinciding with the actual objects on the table. }
%     \label{fig:wromg-grasps}
% \end{figure}

We propose SCARP, a method that performs \textbf{S}hape \textbf{C}ompletion in \textbf{AR}bitrary \textbf{P}oses. Unlike existing methods that have to directly learn a prior over all possible poses and shapes, we first disentangle the pose from the shape of a partial pointcloud.
We build a multi-task objective that: 
(1) generates a disentangled feature representation of pose and shape by canonicalizing an object to a fixed frame of reference, (2)  estimates the exact pose of the object, and (3) completes the shape of the object using the disentangled representation. This multi-task objective allows our network to jointly understand the pose and shape of the input.
% making it robust to different perturbations in the poses of different shape instances. 
%us to perform the different tasks simultaneously making the network robust to the different perturbations in the pose. 
%\Srinath{Why is it better to use a multi-task objective? What are the benefits?}
% Our method builds 
It does so by learning rotationally-equivariant and translationally-invariant pose features using Tensor Field Networks~\cite{modified-tfn}, and global geometric shape features using PointNet++~\cite{pointnet2}.
%and these features are non-linearly combined to perform shape completion.
% These features are first non-linearly combined to recover the full pointcloud using a decoder. Further, the pose features are used to compute the pose of the partial observation. 

\textbf{Application:} Robotic grasp pose estimation~\cite{contact-graspnet, 6dof, dexnet, sameobjectdifferentgrasps} is a challenging area of research that often expects a faithful reconstruction of the scene in 3D. 
% Robotic manipulation is an challenging area of research in robotics involving sub-tasks such as 6D object pose estimation, grasp pose detection, task and motion planning, collision avoidance, etc. Many of these sub-tasks are directly dependent on a faithful reconstruction of the scene in 3D. 
% For instance, SOTA grasp pose estimation networks~\cite{contact-graspnet, 6dof, dexnet, sameobjectdifferentgrasps} expect a 3D scene pointcloud as an input for their networks. 
As shown in Fig.~\ref{fig:banner} (b), under a partial observation, ~\cite{contact-graspnet} generates grasp proposals that directly collide with the actual object in the scene (shown in red). As a result, the manipulator is likely to collide with the object as it attempts to grasp the objects using one of these predicted grasp poses. 
We use SCARP to complete these partial shapes directly in their observed poses and estimate grasp proposals on these completed shapes. We show that SCARP reduces such invalid grasps by $71.2\%$ over predicting grasp poses directly on the partial observations. 
% In this work, we show that shape completion from arbitrary poses results in improved robotic grasping on a tabletop (Fig.~\ref{fig:banner} (d)).
% We propose a new metric called ``GraspError".
% This metric uses a state-of-the-art grasp-proposal network called Contact-Graspnet~\cite{contact-graspnet} to obtain grasps on (1) partial pointclouds and (2) completed pointclouds using SCARP.
% We specifically check for collisions and ``false" grasp proposals. 
% This helps us in determining if the completed pointcloud improves grasping leading to an improvement in tabletop grasping. 
%
% To the best of our knowledge, we are the first to perform pointcloud completion in arbitrary poses. Thus, we compare against a multi-stage pipeline combining \cite{condor}, a SOTA canonicalization network, with the existing SOTA shape completion networks~\cite{sinv, pointr}.
%
To summarize, our contributions are:
\begin{enumerate} 
    \item We propose SCARP, a novel architecture to perform shape completion from partial pointclouds in arbitrary poses. To the best of our knowledge, this is the first work to do so. 
    % \item We propose SCARP, a method \Madhav{a novel Network Architecture or a novel pipeline/framework instead of a method?} to perform shape completion from partial pointclouds in arbitrary poses. \Madhav{which to the best of our knowledge is the first such approach to do so} 
    \item We show for the first time how a multi-task objective can support: (1) canonicalization, (2) 6D pose estimation, and (3) shape completion on partial pointclouds. 
    \item We demonstrate that SCARP outperforms the existing shape completion baselines (with pre-canonicalization) by $45\%$ and improves grasp pose estimation by reducing invalid grasp poses by $71\%$.
    
    % We demonstrate that SCARP outperforms existing baselines (with pre-canonicalization) by $45\%$ 
    % that need an external canonicalization on shape completion in arbitrary poses by $X\%$ 
    % and evaluate the completed output by directly measuring an improvement in grasp proposals for partial pointclouds.
    %, an important task in Robot Manipulations. 
    % the performance improvement on grasp proposals resulting in an improvement in manipulation. 
\end{enumerate}

\section{Related Work}

\textbf{Partial Pointcloud Completion} has been extensively addressed over the years~\cite{pointr, pcn, mvp, crn, grnet, shapeformer, pcusingshapeprior, softpoolnet, seedformer, selfsupcrn, sinv, song2016ssc}. Early 3D shape completion works relied on intermediate voxel representation for representing the 3D objects~\cite{voxel1, voxel2, voxel3}. More recent works adopted an architecture similar to PointNet~\cite{pointnet} that aggregated point-wise embeddings to achieve a global pointcloud feature. 
%applied MLPs independently on each point and aggregated these features to achieve a global geometric feature. 
PCN~\cite{pcn} adopted such an architecture and pioneered learning-based methods for pointcloud completion. More recent works~\cite{sinv, shapeformer} adopted a pointcloud generator network to learn a prior over the full pointcloud shapes. ~\cite{sinv} applied a  differential degradation layer on the output of a pointcloud generator~\cite{tree-gan} to obtain partial pointclouds from full outputs. Later, it relied on conditional GAN inversion~\cite{ganinversionsurvey} to obtain the latent code of a partial pointcloud's full output.
%it used the technique of GAN inversion to obtain the latent code of the corresponding full pointcloud and used the generator to generate the full pointcloud back. 
~\cite{shapeformer} used a similar approach by training a quantized auto-encoder and adopting a second transformer based network to learn the shape prior over the quantized space. Other lines of work aim to generate high resolution pointclouds~\cite{pointr, seedformer}, preserving high-details in the output. All of these existing methods expect the partial pointcloud to be in a fixed canonical frame. As a result, they depend on external canonicalization methods~\cite{condor, capsule_networks, nocs} to bring the partial inputs to their fixed frame. 
%with zero rotation and positioned such that its corresponding full pointcloud is mean-centered at zero. 
% Since in such a space, a pointcloud in any new pose is a new pointcloud, training these networks directly on random rotations results in unnatural pointclouds. We use a similar approach of learning a prior over the full shape using a auto-encoding network. 
%Unlike the existing methods, 
We explicitly disentangle the pointcloud's pose and shape and train our network to learn a prior over the pose and the shape separately. This allows us to perform partial pointcloud completion directly in any arbitrary pose. 
%We specifically focus on completion for improved tabletop manipulation. Thus, we focus on proper canonicalization and improved grasp proposals. 
\cite{weaklysup, mvp} aim to complete pointclouds in arbitrary poses but need multiple observations from multiple views for completing the pointcloud. Capturing a particular object from multiple different views is not possible always, especially in a moving frame of reference like moving cars. 
%We complete partial pointclouds in arbitrary poses observed from a single observation. 
\cite{limitedposevariation} performs pointcloud completion with limited disturbance from the canonical pose and uses temporal information for  completion. Unlike them, we do not assume any restriction in the pose and perform pointcloud completion from a single observation.
% !!!!!!! =====> !!!!!!!!
% limited pose variation one?
% !!!!!!! <===== !!!!!!!!

\begin{figure*}[t]
    \centering    \includegraphics[width=\linewidth]{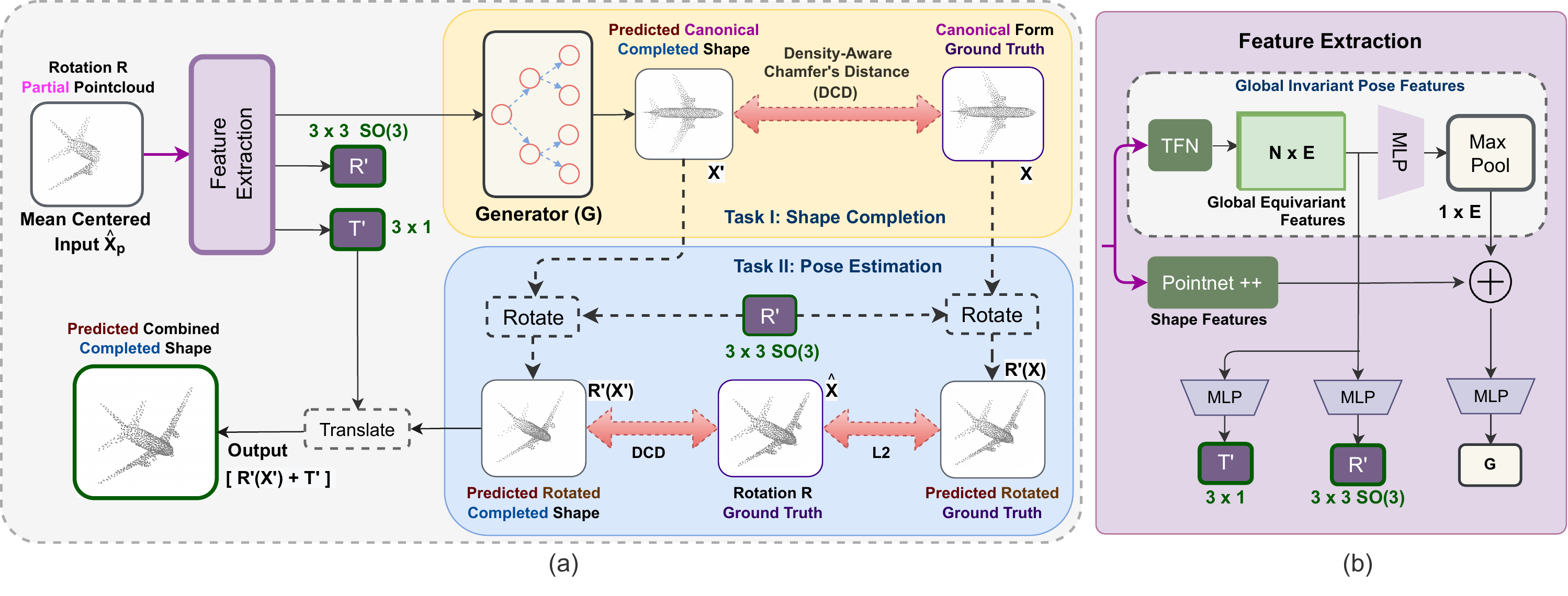} 
    \caption{\small \textbf{Overview of our proposed approach:} The input to SCARP is a mean-centered partial pointcloud $\hat{X_p}$ in an arbitrary orientation $R$. Our feature extraction module \textbf{(b)} disentangles the partial pointcloud's pose and shape and is trained in a multi-tasking objective \textbf{(a)}. In the first task, SCARP combines Pointnet++~\cite{pointnet2} and TFN~\cite{modified-tfn} features to generate a shape feature that is used by a pointcloud completion network, $G$, to generate $X'$. In the second task, the TFN pose feature is used to generate an equivariant frame $\{R', T'\}$. Our loss functions enable the overall network to learn a prior over the shape while understanding the pose of the partial input.}
    % One loss function 
    % \kd{first to what.. no other mention of loss functions other than this in the caption}
    % trains the network to generate the full shape in a canonical form $X'$ against the canonical ground truth $X$. Simultaneously, $R'$ is trained against $R$ by rotating the ground truth canonical pointcloud $R'(X)$, the completed canonical output $R'(X')$ and comparing the rotated pointclouds with $\hat{X}$.} 
    %The final output is a combination of the partial and the completed pointcloud given as $[R'(X') + T'] + \hat{X_p}$. }
    \label{fig:main_architecture}
\end{figure*}

\textbf{Pointcloud Canonicalization} canonicalizes an input pointcloud of a given category to the category's fixed canonical frame. This canonical frame is implicitly defined by the network~\cite{condor, capsule_networks}.  Such a method can be clubbed with the existing shape completion networks 
%that expect the pointclouds in a fixed canonical frame 
in a multi-stage pipeline: (1) canonicalizing the partial pointcloud to a fixed implicit frame, (2) shape completion in the fixed implicit frame, and (3) de-canonicalization. However, the performance of the shape completion network is tightly coupled with the canonicalization method. Morever, there is an additional training overhead in such a method, where the existing dataset has to be first converted to the implicit canonical frame before the shape completion model is trained. Our model does not need an external canonicalization. We compare our model with the multi-stage pipeline by clubbing \cite{condor} with the existing shape completion networks ~\cite{pointr, sinv} and show substantial performance gain on shape completion metrics.

%However, such a mechanism needs explicit model training on datasets created on the network's canonical frame. 

\section{Background}
\label{sec:background}

\underline{Pointnet++~\cite{pointnet2}}
\label{sec:back-point}
is a 3D hierarchical network, $P$, that computes the geometric shape feature, $p$, for a given pointcloud $X \in \mathbb{R}^{3 \times K}$, where $K$ is the user-defined number of points in the pointcloud. It computes a global geometric feature by hierarchically aggregating local geometric features of the points in a pointcloud. For a more detailed understanding, please refer~\cite{pointnet, pointnet2}.
%At each hierarchy $i$, $M_i$ groups are formed by first sampling $M_i$ farthest points as the centroids of the groups and adding additional $Z_i$ points in each group within a predefined radius $R_i$ from the centroid \kd{long sentence hard to follow}. 
% Local geometric features are then computed by aggregating the pointwise features of all the points in each group at each heirarchy $\in \mathbb{R}^{M_i \times E}$. In the final layer $M_L$, the local geometric features of all the remaining centroids $\in \mathbb{R}^{M_{L-1} \times E}$ are aggregated to a global shape feature, $p \in \mathbb{R}^{E}$. At this point, 
These features are not rotation-aware. That is, for a point cloud $X$ in any new orientation $\hat{R}(X)$, where $\hat{R} \in SO(3)$, a unique $p$ is generated. 
% Thus, we use rotationally equivariant TFN embeddings~\cite{modified-tfn} to understand the pose of the pointcloud and rely on $p$ to understand the structure of the  input pointcloud.

\underline{Tensor Field Networks~\cite{org-tfn, modified-tfn}} is a 3D architecture, $\mathcal{X}$, that computes rotationally equivariant and translationally invariant feature matrix, $\hat{F}$. For a given pointcloud $X \in \mathbb{R}^{3\times K}$ and an integer (aka type) $\ell \in \mathbb{N}$, a TFN produces global (type $\ell$) feature vectors of dimension $2\ell + 1$ stacked in a matrix $\hat{F}^\ell \in \mathbb{R}^{(2\ell + 1) \times N}$, where $N$ is user-defined number of channel. $\hat{F}_{:, j}(X)$ satisfies the equivariance property $\hat{F}_{:,j}(RX) = D(R)\hat{F}_{:,j}(X)$, where $D: SO(3) \rightarrow SO(2\ell+1)$ is a Wigner matrix (of type $\ell$)~\cite{wigner1, wigner2, wigner3}. Additional details can be found in \cite{modified-tfn, tfnref1, tfnref2, tfnref3, org-tfn}. 

\underline{tree-GAN~\cite{tree-gan}} is a GAN-based pointcloud generation network that adopts a graph convolutional generator, $G$, and a discrimintor similar to r-GAN~\cite{rgan}. $G$ hierarchically upsamples a gaussian noise, $z \in \mathbb{R}^{E}$, sampled from a standard normal distribution to a pointcloud $X' \in \mathbb{R}^{3 \times K}$.

\underline{Density Aware Chamfer's Distance~\cite{dcd}} Given two pointclouds $X$ and $Y \in \mathbb{R}^{3 \times K}$ without known point-wise correspondences, Chamfer's Distance (CD)\footnote{\textcolor{blue}{\href{https://pdal.io/en/stable/apps/chamfer.html}{https://pdal.io/en/stable/apps/chamfer.html}}} can be used to compute the distance $d_{CD}(X, Y)$ by considering the nearest neighbor of the points $\{x \in X, y \in Y\}$ in $\{Y, X\}$ as their correspondence. 
The distance is then given as:
\begin{equation}
    \sum_{x\in X} \min_{y \in Y} {|| x - y ||}_2^2 + \sum_{y\in Y} \min_{x \in X} {|| x - y ||}_2^2
\end{equation}
% $CD$ can also be used as an optimization objective to minimize the distance between a predicted pointcloud $X'$ and a ground truth pointcloud $X$. 
However, CD does not guarantee uniformity in the output density.
% The predictions are expected to minimize an average objective resulting in points accumulated non-uniformly at certain locations in the 3D space. 
Density-Aware Chamfer's Distance (DCD)~\cite{dcd} overcomes this issue by modifying CD as:
% $d_{CD}(X,Y)$ to $d_{DCD}(X, Y)$ as:
\begin{equation}
    \frac{1}{2}
        \left(
            \frac{1}{|X|} \sum_{x \in X}
            \left(1  - \frac{e^{\mathcal{Z}_x}}{n_{\hat{y}}} \right)
             + \frac{1}{|Y|} \sum_{y \in Y}
                \left(1 -         \frac{e^{\mathcal{Z}_y}}{n_{\hat{x}}}
                \right)
        \right)
\end{equation}
Please refer to \cite{dcd} for the exact notations.

\section{SCARP: Shape Completion in ARbitrary Poses}

Given a partial object pointcloud $\hat{X_p}$ at an unknown pose $\{R, T\}$, we want to estimate this pose and the corresponding full object pointcloud $\hat{X}$ in the same pose. 
%Given a partial pointcloud $X_p$ at a pose $R(X_p) + T$, we want to estimate the corresponding full pointcloud in the same pose $R(X) + T$ and the pose $\{R, T\}$ 
% Here, $\hat{X}$ is a single object.

This is a challenging task as for a neural network, a pointcloud $X$ in two different poses $\{R_1, T_1\}$ and $\{R_2, T_2\}$ are two completely different pointclouds. Thus, we adopt a multi-tasking objective that disentangles the pose and the shape of the input partial pointcloud $\hat{X_p}$. 
The shape component allows us to understand that $\hat{X_p}$ is a partial observation of $X$ which is $\hat{X}$ in its canonical form. The pose component is then used to estimate the pose transform $\{R, T\}$ between $\hat{X}$ and $X$.
% The shape component of this disentangled representation allows our network understand that $\hat{X_p}$ is a partial observation of a an object $X$ which is $\hat{X}$ at its canonical form. The pose component allows our network realize the pose $\{R, T\}$ transform between $\hat{X}$ and $X$.

%$X_{p\{R_1, T_1\}}$ and $X_{p\{R_2, T_2\}}$ are the partial observation of a single object $X$ in the two different poses $\{R_1, T_1\}$ and $\{R_2, T_2\}$. The overall architecture of our network is shown in Fig.~\ref{fig:main_architecture}.

\subsection{Multi-tasking Pipeline for disentangling Shape and Pose}

Let $X_p$ and $X$ be a partial and its corresponding full pointcloud in a fixed canonical frame. Then $\hat{X_p}$ and $\hat{X}$ are $X_p$ and $X$ in an \underline{unknown} arbitrary pose $\{R, T\}$ such that $\hat{X_p} = R(X_p) + T$ and $\hat{X} = R(X) + T$. 
The input to our network is
%a partial pointcloud 
$\hat{X_p}$ which is 
%in an unknown orientation 
mean centered at the origin. Our aim is to predict $\{R, T\}$ and the full pointcloud $\hat{X}$ which is posed as:
%and $R(X_p) + T = \hat{X_p}$.
% $R$ and $T$ transforms $X$, to $\hat{X}$. 
% We pose this as:
\begin{equation}
    \{R, T, \hat{X}\} = \Phi(\hat{X_p})
\end{equation}
where $\Phi$ denotes our proposed network, SCARP.

% and a rotation component $R$ given as $R(X_p)$. Our aim is to predict $R$ and $T$, where $T$ is the mean of $X_p$ in $X$ if $X$ is mean-centered at $0$, along with the full pointcloud $R(X)$. We pose this as:
% \begin{equation}
%     \{R, T, R(X)\} = \Phi(R(X_p))
% \end{equation}
% where $\Phi$ denotes our proposed network, ArPCom.

% As every new rotation $\bar{R} \in SO(3)$ on $X$ is a unique pointcloud for a neural network, learning to estimate the complete shape of every new partial instance in any new orientation in the $SO(3)$ space will cause the underlying space to explode. To tackle this, we formulate a 
Our multi-tasking objective is formulated to % to (1) predict the completed shape $X$ in its canonical form, (2)
(1) complete the partial pointcloud in a fixed canonical frame given by $X$ and (2) estimate the pose transformation from the canonical frame to the original pose $\{R, T\}$. In this pipeline, the two components (1) pose and (2) shape are predicted separately using two different output heads as shown in Fig.~\ref{fig:main_architecture}. 

\subsubsection{Feature Extraction} 

To estimate the input's shape, we compute global geometric shape features, $p \in \mathbb{R}^E$, using Pointnet++~\cite{pointnet2} as explained in Sec.~\ref{sec:background}.
To estimate the pose of the input, we adapt TFN~\cite{modified-tfn} as explained in Sec.~\ref{sec:background}. %to compute 
%the pose features.
%compute invariant pose embeddings and an equivarient coordinate frame. 
Our TFN computes a global equivariant feature, $F \in \mathbb{R}^{N \times E}$ by max pooling over the types $\{\ell\}_{\ell = 0}^{\ell=\ell_{max}}$, where $E$ is the dimension of the equivariant embeddings, $N$ and $\ell_{max}$ are user-defined. 

The input to our shape completion network is a non-linear combination of $p$ and a global invariant embedding, $F_\mathcal{X} \in \mathbb{R}^E$, computed by max pooling $F$ over the channel dimension, $N$. Additionally, $F$ is used to estimate an equivariant frame of reference, $\{R' \in \mathbb{R}^{3 \times 3}, T' \in \mathbb{R}^3\}$ that transforms the invariant embeddings to $X$'s original pose.

\subsubsection{Task I: Shape Completion}

% \begin{figure}[t]
%     \centering    \includegraphics[width=0.7\linewidth]{images/correspondences._V3.pdf} 
%     \caption{\small Chamfer's Distance is negligible as the object is flipped about the XY-axis even though pointwise correspondences are far.}
%     \label{fig:flipped-axis}
% \end{figure}

Completing the shape of a partial input at any arbitrary orientation 
%($\hat{X_p} = R(X_p)$, where $R$ is unknown) 
is difficult. Therefore, we aim to first complete the shape at a fixed canonical frame. To learn this canonical frame, the model needs to build an understanding of the full shape of the partial input. To achieve this, we train our model to predict 
%Learning this canonical form is difficult if the model does not understand the underlying shape of the partial input. Thus, we train our model to perform shape completion and canonicalization simulateneously by predicting 
%by first canonicalizing the input to a fixed frame. 
% However, canonicalizing a partial input is challenging if the model does not understand the underlying shape of the partial input. 
% Thus, our model is trained to predict 
a full canonicalized pointcloud $X'$ 
%as a function of $\hat{X_p}$.
%directly in a fixed canonical frame. $X'$
% that captures the shape of $X$ 
directly from $\hat{X_p}$. Shape completion enables our model to learn a prior over the global shape of a category (a typical chair would have four legs and a backrest) enabling our network to directly canonicalize the partial inputs accurately.  
% Shape completion allows the model to learn a prior over the global shape of an input object (a typical chair would have four legs and a backrest) allowing the network to perform canonicalization on direct partial inputs. 

We adopt $G$, as explained in Sec.~\ref{sec:background}, as our shape completion network where (1) the input to $G$ is a semantically meaningful embedding generated from a partial input $\hat{X_p}$ and (2) is trained using a distance loss against the full pointcloud $X$ to learn a relationship between the partial input $\hat{X_p}$ and the predicted full canonical pointcloud $X'$.   
%We adopt tree-GAN~\cite{tree-GAN}, denoted as $G$, as our decoder. 
As shown in Fig.~\ref{fig:main_architecture} (right), the input to $G$ is a globally invariant feature vector $f \in \mathbb{R}^{E}$ computed by combining $p = P(\hat{X_p})$ and $F_\mathcal{X} = \mathcal{X}(\hat{X_p})$ non-linearly using a neural network $\phi_S$ given as:
\begin{equation}
     X' = G(f)\quad \textrm{and}\quad f = \phi_S(\mathcal{X}(\hat{X_p})\oplus P(\hat{X_p})))
\end{equation}
% \begin{equation}
%     X' = G(f)
% \end{equation}

\subsubsection{Task II: Pose Estimation}

Once $G$ predicts the full pointcloud $X'$ in a canonical pose, it is important to estimate the correct rotation $SO(3)$ matrix $R \in \mathbb{R}^{3\times3}$ and translation $T \in \mathbb{R}^{3}$ to register $X'$ back on $\hat{X_p}$. We predict $R'$ and $T'$ on the second head of our model using the rotationally equivariant TFN features $F$ given as:
\begin{equation}
    R' = \phi_R(F)\quad \textrm{and}\quad T' = \phi_T(F)
    \label{eqn:randt}
\end{equation}
where $\phi_R$ and $\phi_T$ are multi-layered perceptrons.

\subsection{Loss Functions for Multitask Training}

\subsubsection{Shape Completion in a fixed Canonical Frame}

In the first task, we estimate the completed pointcloud in a fixed canonical frame given by $X'$. 
We use DCD (see Sec.~\ref{sec:background}) to minimize the distance between the predicted pointcloud $X'$ and the ground truth canonical pointcloud $X$ given by:
\begin{equation}
    \mathcal{L}_{\mathit{shape}} = d_{DCD} (X', X)
\end{equation}
% Training $G$ over many instances of a given category allows it to learn a prior over a shape in a fixed canonical frame. 

\subsubsection{Estimating the pose of the object} 

\begin{figure*}[t]
    \centering
    \includegraphics[width=0.9\linewidth]{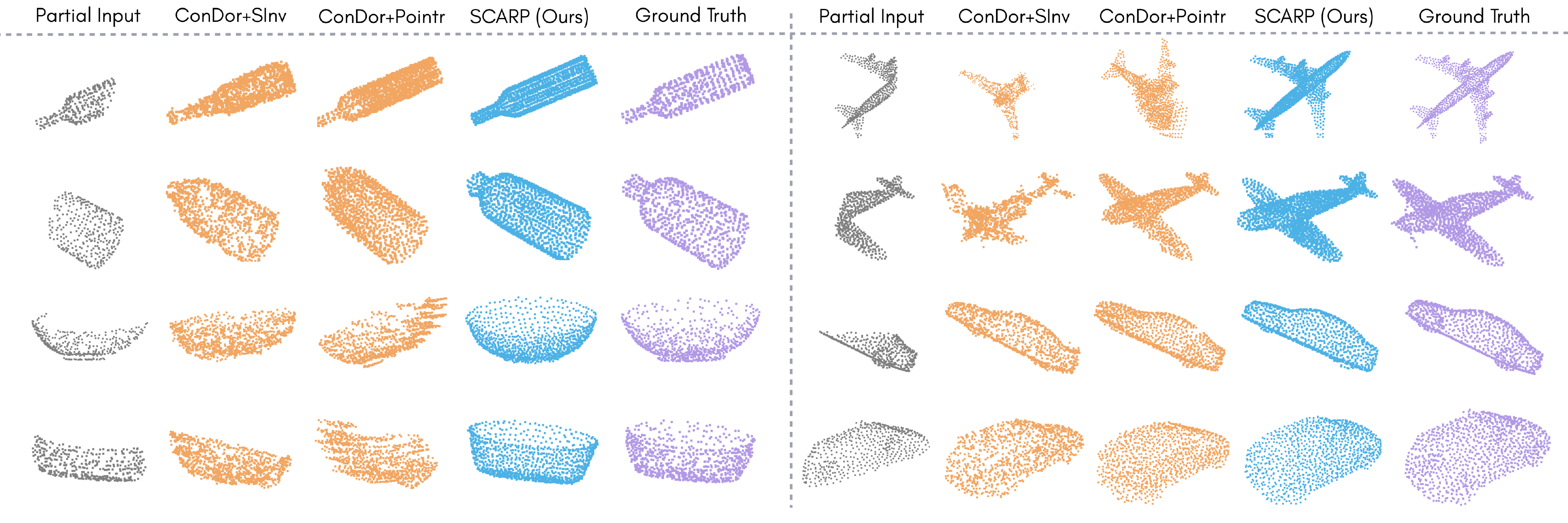}
    \caption{\small Qualitative comparison of shape completion in arbitrary poses on SCARP and the existing multi-stage baselines: Canonicalization using ConDor, Shape Completion, and De-canonicalization.  Pointr\cite{pointr} is a SOTA pointcloud completion network that generates high-resolution completed pointclouds. Shape Inversion (SInv.)~\cite{sinv} is based on tree-GAN~\cite{tree-gan} that shares our generator $G$. }
    \label{fig:qualitative}
\end{figure*}

\begingroup
\begin{table*}[t]
    \centering
    % \caption{Caption}
    % \label{tab:my_label}
    % \rowcolors{1}{gray!25}{white}
    \adjustbox{max width=\textwidth}{
    \begin{tabular}{|c|r|ccccc|cccc|c|}
        \hline
       & & \multicolumn{5}{c|}{Tabletop} & \multicolumn{4}{c|}{Off-Table} & \\
        \hline
        \rowcolor{gray!25}
        & & Bowl & Bottle & Can & Mug & Basket & Plane & Car & Chair & Watercraft & Average
         \\
         \hline
         \multirow{3}{*}{CD$\downarrow$}  & ConDor$+$SInv. & 82.7 & 27.4 & 45.4 & 41.5 & 85.3 & 34.2 & 14.7 & 59.4 & 39.9 & 47.8 \\
        & ConDor$+$Pointr & 30.8 & 20.9 & 29.9 & 14.2 & 40.9 & 22.1 & 6.4 & 19.8 & 8.5 & 21.5 \\
        & SCARP (Ours) & \textbf{21.8} & \textbf{7.9} & \textbf{11.8} & \textbf{12.1} & \textbf{34.2} & \textbf{6.9 } & \textbf{5.6} & \textbf{19.1} & \textbf{7.1} & \textbf{14.0}\\
         \hline
         \rowcolor{gray!25}
         \multirow{5}{*}{MMD-EMD$\downarrow$} & & Bowl & Bottle & Can & Mug & Basket & Plane & Car & Chair & Watercraft & Average
         \\
         \hline
        %  PCN~\cite{pcn} &  \\
         & ConDor$+$SInv. & 27.3 & 17.2 & 20.1 & 19.9 & 29.2 & 19.6 & 11.3 & 22.2 & 18.9 & 20.6 \\
        %  ConDor $+$ FoldingNet~\cite{seedformer} & \\
         & ConDor$+$Pointr & 21.6 & 13.6 & 14.8 & 12.6 & 18.8 & 14.4 & 8.1 & 13.5 & 9.1 & 14.1 \\
         & SCARP (Ours) & \textbf{9.6} & \textbf{6.3} & \textbf{8.8} & \textbf{8.4} & \textbf{10.6} & \textbf{5.0} & \textbf{5.6} & \textbf{8.4} & \textbf{6.0} & \textbf{7.6} \\
         \hline
    \end{tabular}}
    \caption{\small Quantitative comparison of shape completion in arbitrary poses for tabletop and off-tabletop  objects. Most tabletop objects are symmetrical whereas off-table objects have more variations in structure. Chamfer's Distance (CD) and Earth Movers Distance-Maximum Mean Discrepancy (MMD-EMD) are explained in Sec.~\ref{sec:experiments} and are scaled by $10^3$ and $10^2$.}
    \label{tab:main_results}
\end{table*}
\endgroup
% CD scaled by 1000
% BOWL: 3800
% MUG: 4300
% CAN: 2200
% BOTTLE: 5000
% PLANE: 800 * 5 (6.8)

To estimate the pose given by $\{R, T\}$, we use rotationally equivariant pose features $F$ and pass it through $\{\phi_R, \phi_T\}$. We constrain this prediction against the canonical frame. To do so, we rotate the canonical output $X'$ to obtain $R'(X')$ and compare it against the rotated ground truth $\hat{X}$. At this point however, the pointwise correspondences between $X$ and $X'$ are lost. Thus, a hard distance loss such as Euclidean distance cannot be directly used. To tackle this, we minimize permutation invariant CD objective as explained in Sec.~\ref{sec:background} between $X$ and $X'$. However, CD only minimizes the distance between the nearest neighbors of the points in the pointcloud. This results in local minimas where the loss is minimal even when the actual correspondences are far.
%as shown in Fig.~\ref{fig:flipped-axis}. 
As a result, the predicted pointcloud is often flipped about one of the axes. To tackle this issue, we rotate the canonical ground truth $X$ using the predicted $R'$ and compare against $\hat{X}$ using $L2$ loss. The overall loss is: % given as: 
\begin{equation}
    \mathcal{L}_{\mathit{rot}} = \delta d_{CD}(\hat{X}, R'(X')) + \gamma {|| \hat{X}, R'(X)||}_2
\end{equation}
$R'(X')$ is computed by detaching the forward computation graph at the output of $G$. The gradients from the loss does not backpropogate through $G$ at the first head.

For symmetrical objects such as bowls and glasses, multiple $R'$ predictions can be correct.
%(Fig.~\ref{fig:flipped-axis}, right). 
A hard $L2$ loss penalizes the network for correct predictions even for correct $R'$ if the correspondences do not exactly match. Thus, for symmetrical objects, we keep $\delta \sim 1.0$ and $\gamma \sim 0.0$ and for non-symmetrical objects we keep $\delta \sim 1.0$ and $\gamma \sim 1.0$. 

The input to our network is a mean-centered partial pointcloud $\hat{X_p}$. At this point, we train our network to regress to $\hat{X_p}$'s centroid in the full pointcloud $X$ given by $T'$. We directly supervise $T'$ against the ground truth $T$ given as: 
\begin{equation}
    \mathcal{L}_{\mathit{trans}} = {||T' -  T||}_2
\end{equation}

The final output is obtained by rotating and translating our predicted pointcloud $X'$ by $R'$ and $T'$ respectively as: 
%as $R'(X')$, translating the pointcloud by $T'$ as:
%, and finally combining it with the partial input $R(X_p)$ as:
\begin{equation}
    X_o = R'(X') + T' 
\end{equation}
% As a postprocessing step, we merge $X_o$ and $\hat{X_p}$ allowing us to preserve the details in the partial input. 
%Combining the output of our network with the partial input allows us to preserve the details of the partial pointcloud.

\textbf{Orthonormality Loss:} The rotation $R'$ predicted by our network is a $3\times3$ matrix in the $SO(3)$ space. However, the matrix predicted by Eqn.~\ref{eqn:randt} is not guaranteed to be a valid $SO(3)$ matrix. 
We therefore, enforce orthonormality on $R'$ by minimizing its difference to its closest orthonormal matrix. To do so, we compute the SVD decomposition of $R = U\Sigma V^T$ and enforce unit eigenvalues as: 
\begin{equation}
    \mathcal{L}_\mathit{orth} = ||UV^T - R||_2
\end{equation}

\subsubsection{Combined Loss}

We train our network end-to-end by combining all the losses as: 
\begin{equation}
    \mathcal{L} = \mathcal{L}_{\mathit{shape}} + \mathcal{L}_{\mathit{rot}} + \mathcal{L}_{\mathit{trans}} + \mathcal{L}_{\mathit{orth}}
\end{equation}

\section{Experiments}
\label{sec:experiments}

In this section, we evaluate SCARP on two tasks: \textbf{(T1)} Shape completion in arbitrary poses and \textbf{(T2)} Improving grasp proposals by completing partial pointclouds.
% In this section, we evaluate SCARP on \textbf{(T1)} Shape completion in arbitrary poses and \textbf{(T2)} Improvement in grasp pose estimation for partial pointclouds. 
% \Srinath{What are T1 and T2 supposed to mean? Experiments? Tasks?}
%Improving Grasp proposals on partial pointclouds.
%\cite{contact-graspnet}.

\textbf{Baselines:} As we are the first to perform the task T1, we modify the existing shape completion networks by developing a multi-stage pipeline: (1) We use ConDor\cite{condor} to first canonicalize the input partial pointclouds to a fixed canonical frame defined implicitly by ConDor. (2) We train and test the existing shape completion methods on ConDor's canonical frame. (3) Bring the completed pointcloud to the original orientation using a pose transform predicted by ConDor. We compare against (1) ConDor$+$Pointr\cite{pointr}, a SOTA pointcloud completion network that generates high-resolution completed pointclouds and (2) ConDor$+$Shape Inversion (SInv.)~\cite{sinv} based on tree-GAN~\cite{tree-gan} that shares our generator $G$.
% , $G$, the generator used by SCARP for shape completion.% , the generator of our network.
%the generator ($G$) of our network.

\textbf{Metrics: } We use Chamfer's Distance \textbf{(CD)} as explained in Sec.~\ref{sec:background} to compute the distance between the ground truth pointcloud $\hat{X}$ and the predicted pointcloud given as $R'(X') + T'$ to evaluate the match in shape. 
%Our primary metric is Chamfer's Distance \textbf{(CD)}. As explained in Sec.~\ref{sec:background} CD computes the distance between the ground truth $\hat{X}$ and the predicted pointcloud given as $R'(X') + T'$ to evaluate the match in shape. 

Earth Movers Distance-Maximum Mean Discrepancy \textbf{(MMD-EMD)}~\cite{tree-gan, rgan} is used to evaluate for uniformity in the prediction by conducting bijective matching of points between two pointclouds. As we only want to measure the output's uniformity, we compute this metric between the canonical ground truth $X$ and the canonical prediction $X'$.
% As we measure the uniformity in the output in this metric, we compute it on the canonical predicted and ground truth pointclouds, $X'$ and $X$ respectively.
% Uniformity in the resulting pointcloud is important and may be needed for many downstream tasks in Robotics (such as collision detection).
% for Robotics that depend on a valid poincloud reconstruction for many downstream tasks. 
% EMD conducts bijective matching of points between two point clouds to measure uniformity in the output pointcloud. 

\begin{figure}[t]
    \centering
    \includegraphics[width=\linewidth]{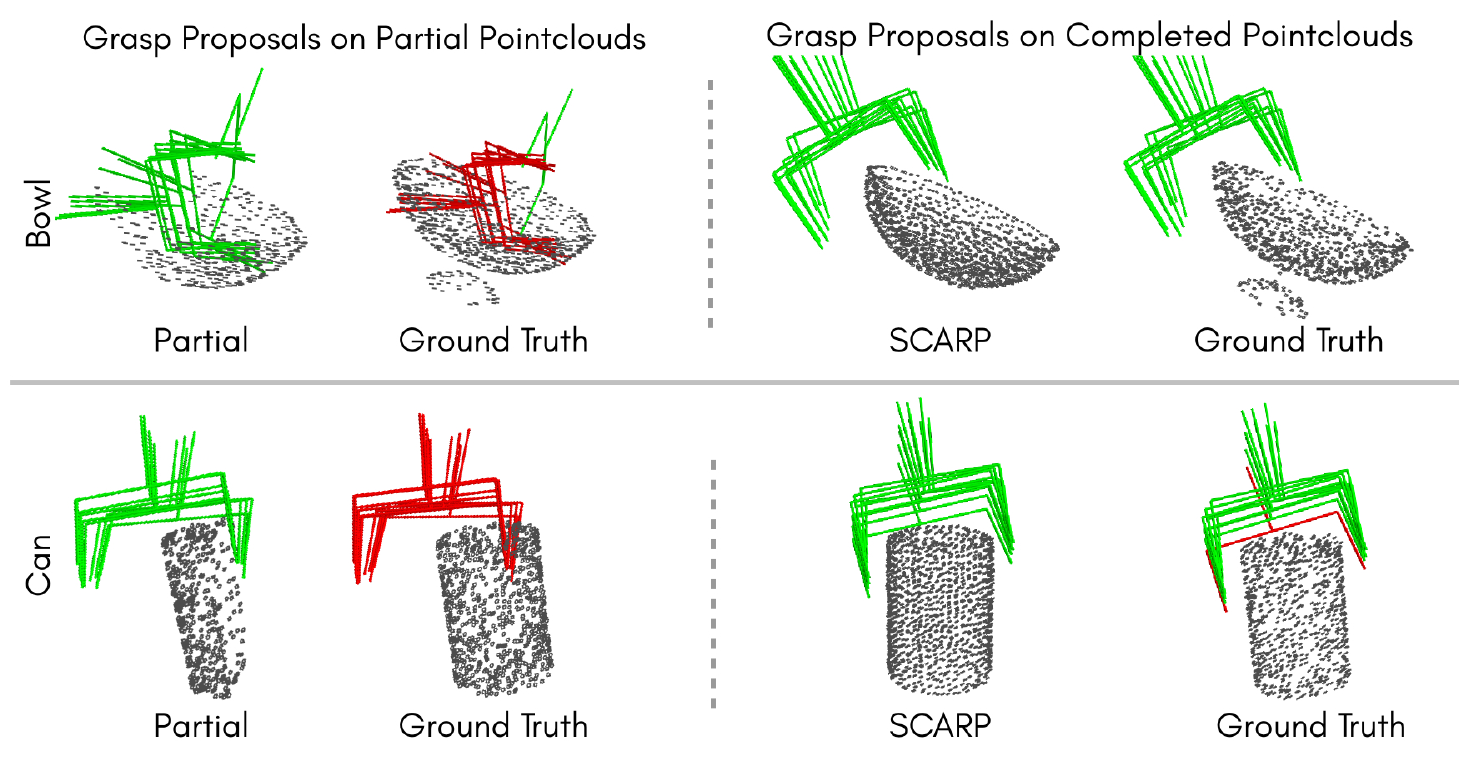}
    \caption{\small \textbf{(left)}: Grasp proposals made by a SOTA grasp proposal network, \cite{contact-graspnet}, on partial observations lead to collisions with the actual object. \textit{Partial} is a partial observation and \textit{Ground truth} denotes the actual object. The proposals are made on \textit{Partial} (shown in green) but collide with the actual object (shown in red). \textbf{(right)}: We use SCARP to complete the partial observations. Grasp proposals made on the completed objects align well with the actual object on the table reducing such collisions by a large margin. 
    % Grasp proposals \cite{contact-graspnet} on (1) partial observations, (2) completed objects using SCARP, and (3) full observations. Red denotes the faulty grasps, Green denotes valid grasps. \kd{hard to say what is 1, 2, and 3 are in the image. Full observations = ground truth? Keep the figure and the caption consistent. Also "completed" = SCARP?}
    % \Srinath{Why is this result interested? What should people look at? Maybe draw a arrow to indicate where people should look at.}
    }
    \label{fig:grasp-qualitative}
\end{figure}

We evaluate SCARP by measuring its impact in an important downstream task: grasp pose estimation. In this, we measure the a) the number of grasp proposals made on the partial object that collide with the actual object on the table (shown in Fig.~\ref{fig:banner} and \ref{fig:grasp-qualitative}) denoted by $C$ and b) number of invalid grasps that do not result in a valid grasping denoted by $I$. We then compute the Grasping Error \textbf{(GE)} as: 
% We propose a novel metric Grasp Error \textbf{(GE)} to evaluate the quality of a shape completion network for the task of grasp proposals on partial poinclouds. As shown in Fig.~\ref{fig:banner} and \ref{fig:grasp-qualitative}, 
% % measure the performance of a shape completion network on grasp pose estimation. As shown in Fig.~\ref{fig:banner}, 
% Contact-Graspnet~\cite{contact-graspnet}, a SOTA grasp proposal network, estimates grasps on the partial object that collides with the actual object (ground truth). We evaluate such collisions on Contact-Graspnet if the partial pointcloud was completed using SCARP \kd{rewrite}. This is measured on  
% % We evaluate the grasp proposals on partial pointclouds after completing these pointclouds using SCARP by measuring 
% % collides with the actual object on the table. 
% % We evaluate the presence of such
% % grasps on the pointcloud completed by SCARP by measuring  
% % recovered full pointcloud by measuring the 
% (A) number of predicted grasps colliding with the actual object on the table, $C$, \kd{what is C here, too many confusing commas} and (B) number of invalid grasp predictions, $I$, outside of the actual object on the table due to incorrect registration of the completed pointcloud \kd{simplify the writing, use a) and b) instead of (A) and (B)}. 
% % that are away from the actual object due to incorrect registration of the recovered pointcloud, $I$. 
% GE is computed as:
\begin{equation}
    \frac{1}{\mathcal{D}}\sum_{i=1}^{\mathcal{D}}\frac{C + I}{\mathcal{N}}
\end{equation}
where $\mathcal{N}$ is the number of top grasp proposals and $\mathcal{D}$ is the total number of pointcloud instances. In our case, $\mathcal{N} = 30$. 
% To compute a) and b), we use the codebase provided by Contact-Graspnet\footnote{\textcolor{blue}{\href{https://github.com/adithyamurali/TaskGrasp}{https://github.com/adithyamurali/TaskGrasp}}}.
% \Srinath{Why is GE a bettr metric compared to others? What is the intuition?}

\textbf{Dataset: }Our dataset is a subset of \cite{shapenet} derived from \cite{condor} and \cite{pcn} made of 5 tabletop (Bowl, Bottle, Can, Mug, Basket) and 4 non-tabletop (Plane, Car, Chair, Watercraft) categories. We evaluate $GE$ only on the tabletop objects. 

% We derive our dataset from \cite{condor} and \cite{pcn} and is a subset of ShapeNet~\cite{shapenet} made of 5 tabletop (Bowl, Bottle, Can, Mug, Basket) and 5 non-table top (Plane, Car, Chair, Waterboard, Lamp) categories. For each category, our test set consists of $\sim$5000 instances of arbitrary instances in arbitrary poses. We evaluate $GE$ specifically for the tabletop objects. 

%The grasp error is computed as $\fract{C + I}{\mathcal{N}}$ where $C$, denote the number of grasp predictions that collide with the actual object on the table predicted due t, $I$ denote 

% We report qualitative results in Fig.~\ref{fig:qualitative} and quantitative metrics in Table~\ref{tab:main_results}. Table~\ref{tab:ablation} presents an ablation on the different components of our network. 

% \subsection{Challenges in a multi-stage Shape Completion Pipeline}

% \subsection{Results}
\textbf{Results: }As shown in Table~\ref{tab:ablation}, SCARP outperforms the existing multi-stage baselines on all the categories on an average by $45\%$. The existing shape completion methods rely on the output of an external canonicalization model that suffer from their own inconsistencies as reported in their paper~\cite{condor}. This results in an error propagation as the input to the shape completion networks are not always in the exact canonical forms. The errors in the input map to a larger error in the output of the networks. This is followed by an error in the transform from the canonical form to the original pose. The resulting output of the multi-stage pipeline suffer from high inconsitensies and sub-optimal outputs. 
%most of the categories on an average by $X\%$. Especially on symmetrical tabletop categories like ``Bowl", ``Bottle", ``Can", ``Mug", and ``Basket", ArPCom outperforms by $X\%$. That is, the existing methods perform poorly on symmetrical objects. \cite{condor} relies on accurate correspondences and uses a distance based metric between the correspondences. This results in the model being penalized for correct $R$ in case of symmetrical objects that has multiple correct $R$. This results in error-propogation where the shape-completion network observes different canonical forms and is trained and inferred on different rotations. Morever, the canonical frame predicted by \cite{condor} has inconsisties as reported in the paper causing the shape completion network to suffer. 
Unlike these networks, our model is trained jointly on  both tasks (canonicalization and shape-completion) using a multi-tasking objective. As we show in the ablations, this objective plays a crucial role in achieving a disentangled representation of shape and pose. 
% together. As we show in the ablations, this multi-tasking objective enables our model to improve on both tasks simultaneously by learning a disentangled feature representation of pose and shape. 
% both the task to improve simultaneously
%. The multi-tasking approach 
% as it allows us learn the pose and shape separately through a disentangled feature representation.
%resulting in an improved performance in both. 
Qualitative results are shown in Fig.~\ref{fig:qualitative} that vividly show the closeness of SCARP's output to the ground truth when compared with others.
% In our case, we use a soft-distance metric that minimizes the distance between the nearest neighbors instead. 

\textit{Improvement in Grasp Proposals}: Generating grasp proposals for partial pointclouds is a challenging task as a network may mistake a missing portion of an object as a potential area to grasp (see Fig.~\ref{fig:banner} and Fig.~\ref{fig:grasp-qualitative}). We apply SCARP to complete these partial observations directly in the observed poses and predict grasp poses on these completed pointclouds using a SOTA grasp generation network Contact-Graspnet~\cite{contact-graspnet}. To evaluate the grasp proposals, we compute GE on (1) partial observations, (2) completed observations by SCARP, and (3) actual objects (ground truth). Actual objects are full pointclouds with no missing portion. As we show in Table~\ref{tab:grasp-quant}, SCARP shows a relative improvement of $71.2\%$ and an absolute improvement of $48.27\%$ over the grasp proposals on the partial pointclouds. Moreover, there is only an absolute degradation of $4.19\%$ vis-a-vis the ground truth. The ground truth error in Table \ref{tab:grasp-quant} is the datum error in the grasp proposals output by Contact-Graspnet.  Qualitative results are shown in Fig.~\ref{fig:grasp-qualitative}. Green and red proposals denote valid and colliding grasp proposals respectively. 
% grasp poses are valid grasp poses, the red grasp poses are colliding poses. 
As can be seen, the grasps proposed on partial observations collide with the actual object (ground truth), whereas, the grasp proposals made on the completed object by SCARP are valid. 
% Fig.~\ref{fig:grasp-qualitative} and Table~\ref{tab:grasp-quant} presents the qualitative and quantitative metrics respectively, for $GE$ \kd{keep the GE or $GE$ font consistent} on (1) partial pointclouds, (2) recovered full poincloud predicted by SCARP, and (3) ground truth full pointcloud. As can be observed, recovering full pointcloud leads to a significant improvement in the proposals made by \cite{contact-graspnet}. This indicates that SCARP can potentially be used to improve various robotic tasks starting from manipulation to autonomous driving and many more \kd{statement not required}. A qualitative real-world example is shown in Fig.~\ref{fig:banner} on YCB~\cite{ycb} objects. 
%We leave the exciting challenge of real-world adaptation from synthetic data to the future work. 

% \subsection{Ablation}
% \label{sec:ablation}

% \begin{table}[t]
%     \centering
%     \rowcolors{1}{gray!25}{white}
%     \caption{Caption}
%     \adjustbox{max width=\linewidth}{
%     \begin{tabular}{|r|cc|cc|cc|}
%         \hline
%          & \multicolumn{2}{c|}{Plane} &
%          \multicolumn{2}{c|}{Bowl} & \multicolumn{2}{c|}{Bottle} \\
%          & CD$\downarrow$ & MMD$\downarrow$ & CD$\downarrow$ & MMD$\downarrow$ & CD$\downarrow$ & MMD$\downarrow$ \\
%          \hline
%         %  PCN~\cite{pcn} &  \\
%          Ours \\
%          \hline
%          w/o TFN & \\
%          w/o Pointnet++ & \\
%          w/o SC & \\
%          \hline
%     \end{tabular}}
%     \label{tab:ablation}
% \end{table}

\begin{table}[t]
    \centering
    % \rowcolors{1}{gray!25}{white}
    \adjustbox{max width=\linewidth}{
    \begin{tabular}{|r|r|c|c|c|}
        \hline
         & & w/o SC & w/o $F_\mathcal{X}$ &
         w/o $p$ \\
        %  & & ${|| R'(X), \hat{X}||}_2$ & $d_{CD}(\hat{X}, R'(X'))$ & $d_{CD}(\hat{X}, R'(X'))$ \\
         \hline
        %  PCN~\cite{pcn} &  \\
         \multirow{ 2}{*}{Plane} & Ours & \textbf{3.8} & \textbf{6.9 }& \textbf{6.9}\\
         & Modified & 112.34 & 135.3 & 111.9\\
         \hline
         \multirow{ 2}{*}{Bowl} & Ours & \textbf{14.9} & \textbf{21.8} & \textbf{21.8} \\
         & Modified & 123.7 & 185.3 & 156.4 \\
         \hline
         \multirow{ 2}{*}{Mug} & Ours & \textbf{10.79} & \textbf{12.1 }&\textbf{ 12.1 }\\
         & Modified & 47.6 & 45.5 & 44.3 \\
         % 46 canon
         \hline
    \end{tabular}}
    \caption{\small Ablation Study: We show the affect of removing different components of our network. In w/o SC, we train SCARP as an auto-encoding network to verify if the model can still learns to canonicalize the input pointcloud. In w/o $F_\mathcal{X}$ and w/o $p$, we remove TFN and pointnet features, respectively, and evaluate the quality of shape completion. Each component plays a crucial role in achieving SCARP as is evident by the drop in metrics. Metrics are scaled by $10^3$.}
    % Performance of SCARP denoted as ``Ours" on removing different components of the network denoted as ``Modified". $d_{CD}$ is scaled by $10^3$.}
    % \Srinath{In the caption, say what are the salient parts of the table that readers should look at. Maybe make the best results bold?}
    \label{tab:ablation}
\end{table}

\begin{table}[]
    \centering
    \rowcolors{1}{gray!25}{white}
    \adjustbox{max width=\linewidth}{
    \begin{tabular}{|r|c|c|c|}
        \hline
         & Partial & SCARP & Ground Truth \\
         \hline
         Bowl & 62.18 & 21.14 & 16.86 \\
         Bottle & 46.5 & 7.35 & 6.32 \\
         Can & 81.33 & 22.33 & 16.0 \\
         Mug & 71.33 & 25.0 & 23.5 \\
         Basket & 77.33 & 21.5 & 13.66 \\ 
         \hline
         Average & 67.73 & 19.46 & 15.27 \\
         \hline
    \end{tabular}}
    \caption{\small Quantitative metrics on GE (explained in Sec.~\ref{sec:experiments}): \% of grasp proposals that are invalid or collide with the actual object on the table when the proposals are made on (1) Partial Observations, (2) Shape Completed objects by SCARP, and (3) Actual Objects on the table (Ground Truth). SCARP reduces invalid and colliding grasp proposals by $71.2\%$ when only partial observations are available by accurately completing the object in the observed pose.
    % Quantitative metrics on Grasp Proposals made by \cite{contact-graspnet}. Grasp Error (GE) is explained in Sec.~\ref{sec:experiments}. Ground Truth denotes the best error that can be achieved.
    % \Srinath{What is the takeaway message from this table? Maybe make the best results bold?}
    }
    \label{tab:grasp-quant}
\end{table}

\textbf{Ablation: }SCARP is trained on a multi-tasking objective to achieve: (1) canonicalization, (2) 6D pose estimation, and (3) shape completion. We evaluate the contribution of the different components in our network in achieving these tasks. 

\textit{(A) Canonicalization without Shape Completion}: Canonicalization involves mapping an input $X$ to its category's fixed canonical frame~\cite{nocs, condor, compass}. Learning a canonical frame for a partial input is challenging as a network may struggle to understand the overall structure of the partial shape.
%needed for correctly estimating a category's canonical form. 
In our model, the structure of a category is correctly learned using the shape completion task. Thus, we analyze if SCARP can canonicalize the partial inputs without performing shape completion. 
To evaluate this, we modify SCARP by training to auto-encode the partial input $\hat{X}_p$ while simultaneously estimating its pose $\{R, T\}$. That is, our generator $G$ generates $X_p$ which is $\hat{X_p}$ in its canonical form and uses $R'$ to rotate $X_p$ back to $\hat{X_p}$. To measure the performance, we compute the CD between $G's$ canonical output and the canonical ground truth. In case of SCARP, this is given as $d_{CD}(X', X)$, and in case of ablation, this is given as $d_{CD}(X_p', X_p)$. As shown in Table.~\ref{tab:ablation} (w/o SC), on average $d_{CD}$ on SCARP is 9.83 whereas when the shape completion aspect is removed, the average distance is 94.54. This indicates that the network does not learn anything meaningful if the task of shape completion is removed from the formulation. 

% SCARP uses a multi-tasking objective to train on three different tasks: (1) canonicalization, (2) pose-estimation, and (3) shape completion. In this section, we evaluate the contribution of the different components of our network on these tasks. We evaluate SCARP on 
%We evaluate affect of the different components in ArPCom on 
% \textit{(A) Canonicalization without Shape Completion:} Canonicalization involves mapping a partial input to a fixed frame of reference. The shape completion task allows the network to learn a prior over the full shapes and thus estimate the proper canonical form. In this ablation, instead of generating a full shape from partial inputs, we train the model in an auto-encoding setting where we want to reconstruct the partial input. In this setting, we train $G$ to generate $X_p$ and use $R'$ to rotate $X_p$ back to $\hat{X_p}$. Without the shape completion prior, the network may not be able to learn a proper canonical form.  
%Training $G$ on pointcloud reconstruction. Shape Completion allows $G$ to learn a prior over the full shape in turn helping the canonicalization on partial inputs. In this ablation, we simply train $G$ to reconstruct the partial intpu $\hat{X_p}$ in a canonical pose. 
\textit{(B) Shape Completion without pose and shape features:} As shown in Fig.~\ref{fig:main_architecture}, $G$ expects a disentangled feature embedding that is a non-linear combination of the pose $F_\mathcal{X}$ and shape embeddings $p$. We remove these features one by one and observe their impact on shape completion. We measure the performance of shape completion as $d_{CD}(\hat{X}, R'(X'))$. As shown in Table~\ref{tab:ablation} (w/o $F_\mathcal{X}$ and w/o $p$), SCARP fails to converge without either of the features. In both cases, the model fails to learn a correct transformation between the canonical and the original pose. Without pointnet (w/o $p$), the model collapses to a few different shapes across the instances missing out on per-instance details. Without TFN (w/o $F_\mathcal{X}$), completion in the canonical frame is more accurate but TFN fails to estimate the correct pose transform. In summary, as the shape completion in the canonical form suffers, the pose transform is also inaccurate thus indicating the importance of the multi-task objective.

\section{Conclusion}

% generative network for Pointcloud Shape Completion in Arbitrary Poses. 
Existing shape completion works assume the partial inputs to be in a fixed canonical frame. This is difficult to achieve in a robotics setting where the objects are observed in arbitrary poses thus needing pre-canonicalization. This leads to an error propagation resulting in a sub-optimal shape completion. We propose SCARP, a novel architecture that performs Shape Completion in ARbitrary Poses. SCARP is trained using a multi-task objective to perform (1) canonicalization, (2) 6D pose estimation, and (3) shape completion. SCARP outperforms the existing multi-stage baselines by $45\%$ and showcases its potential in improving grasp proposals on tabletop objects, reducing colliding grasps by more than $70\%$. SCARP has a huge potential in many more robotics applications like collision avoidance in trajectory planning or differential simulators for model-based RL planners.  

\textbf{Acknowledgement: }We would like to thank Karthik Desingh, who is an Assistant Professor at the University of Minnesota, and Adrien Poulenard, who is a Postdoctoral Fellow at the Stanford University, for their valuable feedback.

%that is difficult for real-world objects and scans. 
% Requiring an explicit canonical frame limits the application of existing shape completion networks to shape completion objects in their arbitrary poses. However, relying on an external pointcloud canonicalization network leads to error-propagation. 
% % Relying on the output of a second unrelated network leads to error-propogation \kd{rewrite}. 
% SCARP is an end-to-end network trained with a multi-tasking objective performing (1) canonicalization, (2) object pose estimation, and (3) shape completion in a single network. We show in the experimental section, SCARP outperforms the multi-stage pipelines of an existing canonicalization network followed by shape completion and improves the performance of grasp proposals on partial pointclouds.
%The multi-tasking objective 

\bibliographystyle{IEEEtran}
\bibliography{eg}

\end{document}